\documentclass{article}

\usepackage{microtype}
\usepackage{graphicx}
\usepackage{subfigure}
\usepackage{booktabs} 
\usepackage{multirow}
\usepackage{tabularx}
\usepackage{placeins} 
\usepackage{balance}  

\usepackage{hyperref}

\usepackage[accepted]{icml2025}

\usepackage{amsmath}
\usepackage{amssymb}
\usepackage{mathtools}
\usepackage{amsthm}

\usepackage[capitalize,noabbrev]{cleveref}

\theoremstyle{plain}

\theoremstyle{definition}

\theoremstyle{remark}

\usepackage[textsize=tiny]{todonotes}

\icmltitlerunning{MedBioLM: Optimizing Medical and Biological QA with Fine-Tuned LLMs and RAG}

\begin{document}

\twocolumn[
\icmltitle{MedBioLM: Optimizing Medical and Biological QA 
with Fine-Tuned Large Language Models and Retrieval-Augmented Generation}



\icmlsetsymbol{equal}{*}

\begin{icmlauthorlist}
\icmlauthor{Seonok Kim}{ }
\end{icmlauthorlist}

\icmlaffiliation{ }{Korea University}

\icmlcorrespondingauthor{Seonok Kim}{sokim0991@korea.ac.kr}

\icmlkeywords{Machine Learning, ICML}

\vskip 0.3in
]



\printAffiliationsAndNotice{} 

\begin{abstract}
Large Language Models (LLMs) have demonstrated impressive capabilities across natural language processing tasks. However, their application to specialized domains such as medicine and biology requires further optimization to ensure factual accuracy, reliability, and contextual depth. We introduce MedBioLM, a domain-adapted biomedical question-answering model designed to enhance both short-form and long-form queries. By integrating fine-tuning and retrieval-augmented generation (RAG), MedBioLM dynamically incorporates domain-specific knowledge, improving reasoning abilities and factual accuracy. To evaluate its effectiveness, we fine-tuned the model on diverse biomedical QA datasets, covering structured multiple-choice assessments and complex clinical reasoning tasks. Fine-tuning significantly improves accuracy on benchmark datasets, while RAG enhances factual consistency. These results highlight the potential of domain-optimized LLMs in advancing biomedical research, medical education, and clinical decision support.\end{abstract}

\medskip

\begin{figure}[ht]
\vskip 0.2in
\begin{center}
\centerline{\includegraphics[width=0.9\columnwidth]
{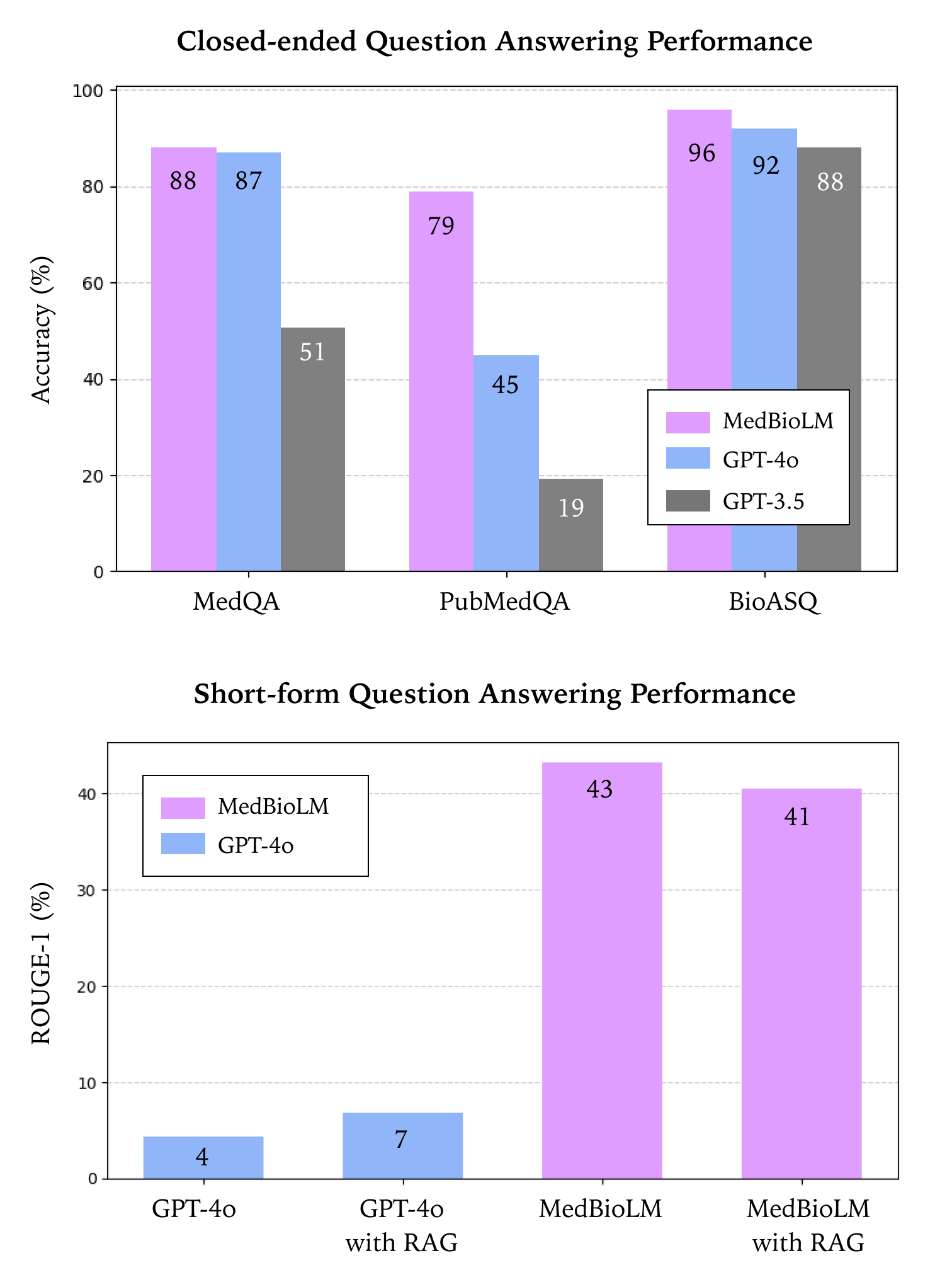}}
\caption{Comparative performance of MedBioLM and base models on closed-ended and short-form biomedical QA tasks, highlighting the benefits of fine-tuning.}
\label{performance_summary}
\end{center}
\vskip -0.2in
\end{figure}

\section{Introduction}
\label{introduction}

The rapid advancements in large language models (LLMs) have significantly transformed their application in specialized domains such as medicine and biology \cite{Oh2023ECGQA,saab2024capabilitiesgeminimodelsmedicine}. These models have demonstrated remarkable capabilities in various question-answering (QA) tasks, ranging from structured multiple-choice reasoning to open-ended long-form explanations \cite{Singhal2023, Luo_2022, saab2024capabilitiesgeminimodelsmedicine, jeong2024olaphimprovingfactualitybiomedical, nori2023generalistfoundationmodelsoutcompete, singhal2022largelanguagemodelsencode, chen2023meditron70bscalingmedicalpretraining}. However, achieving high accuracy and reliability in biomedical QA remains a substantial challenge due to the complexity, domain specificity, and factual accuracy requirements of medical knowledge. Unlike general-purpose QA tasks, medical QA demands a higher degree of precision, interpretability, and contextual depth, making it crucial to explore optimization strategies tailored to this field.

\begin{figure*}[t]
\centering
\includegraphics[width=1.0\textwidth]{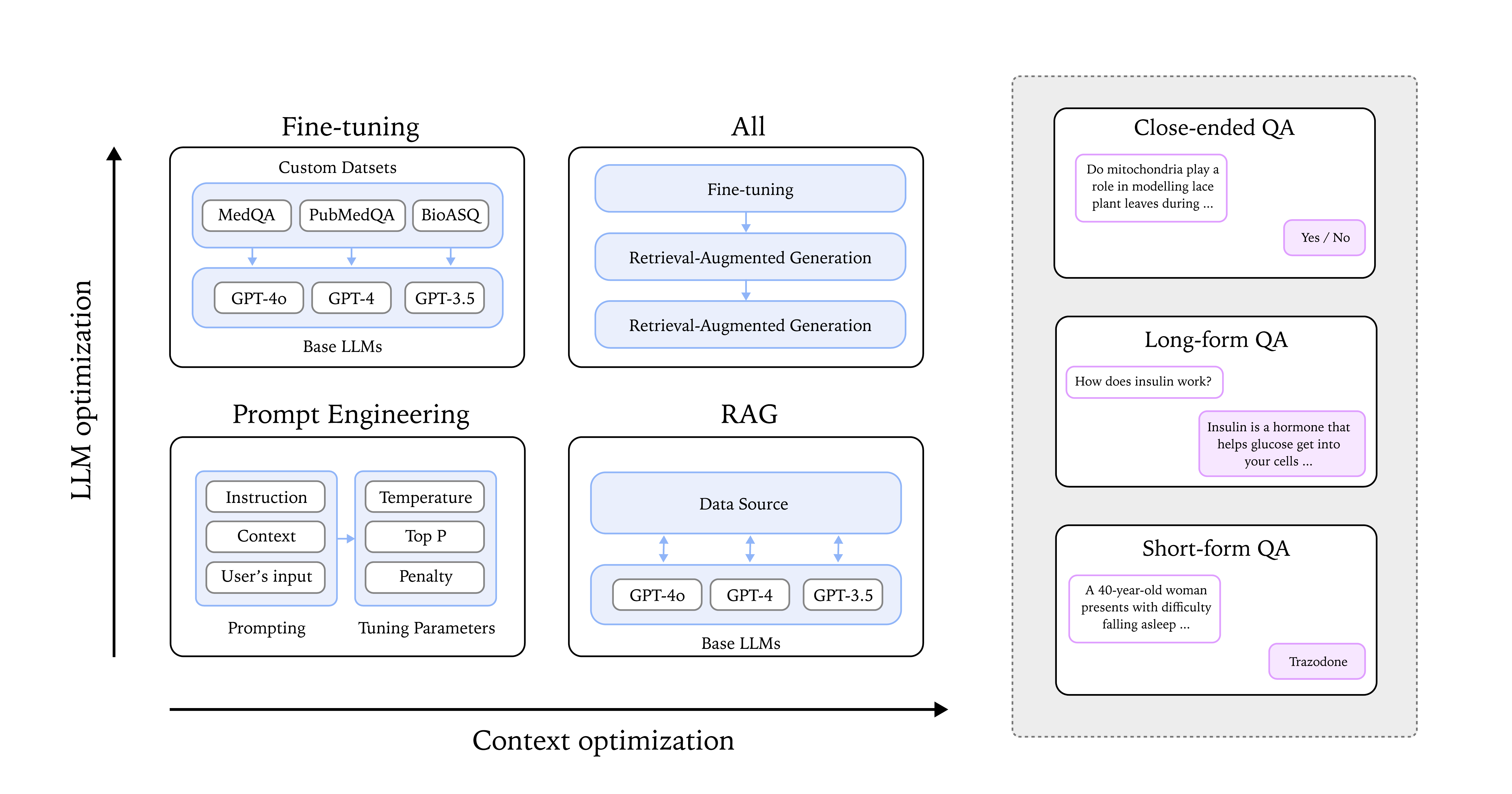}
\caption{Overview of our approach for optimizing large language models (LLMs) in biomedical question answering, integrating fine-tuning, retrieval-augmented generation (RAG), and prompt engineering to enhance performance across different QA formats.}
\label{overview}
\end{figure*}

This study focuses on optimizing LLMs for medical and biological QA tasks by integrating fine-tuning, RAG, and prompt engineering techniques (\cref{overview}). Fine-tuning \cite{fine-tuning} adapts pre-trained LLMs to medical datasets, improving their ability to generate contextually appropriate and factually accurate responses. RAG \cite{RAG} further enhances performance by allowing models to retrieve external domain-specific knowledge, mitigating issues of hallucination and factual inaccuracy. Additionally, prompt engineering plays a crucial role in controlling the response style, ensuring concise and well-structured outputs suited for medical professionals. By systematically evaluating these optimization techniques across multiple QA formats, this study aims to determine the most effective strategy to improve LLMs in biomedical applications.

To assess the impact of these optimization techniques, three primary types of QA tasks were considered:
\begin{itemize}
\item \textbf{Closed-ended QA} (multiple-choice and Boolean reasoning), which requires selecting the correct answer from predefined options, as seen in datasets like MedQA, PubMedQA, and BioASQ \cite{jin2020diseasedoespatienthave, Jin2019PubMedQA, BioASQ, Vilares2019HEADQA}. 
\item \textbf{Long-form QA}, where models must generate detailed explanations based on biomedical literature, requiring coherent, factual, and well-structured responses \cite{BenAbacha2019MedQuAD,LiveMedQA2017, MedicationQA, Jin2019PubMedQA}.
\item \textbf{Short-form QA}, which focuses on concise but precise responses, making it suitable for real-world applications such as clinical decision support and medical search queries \cite{jin2020diseasedoespatienthave}.
\end{itemize}

\cref{performance_summary} presents a comparative analysis of model performance across these tasks. Fine-tuned models demonstrate significant improvements in closed-ended QA, with MedBioLM achieving 88\% accuracy on MedQA and 96\% on BioASQ, outperforming base models like GPT-4o and GPT-3.5 \cite{openai2024gpt4technicalreport, openai2024gpt4ocard}. In long-form QA, the fine-tuned model improves ROUGE-1 scores and BLEU scores compared to GPT-4o on MedicationQA \cite{MedicationQA}. Meanwhile, RAG proves to be highly effective in short-form QA, enhancing factual accuracy and lexical similarity, as reflected in improvement in ROUGE-1 compared to the base model. These results suggest that domain-specific fine-tuning significantly improves structured reasoning, while RAG contributes to more relevant and factual responses in retrieval-dependent queries.

Our contributions can be summarized as follows:
\begin{itemize}
\item RAG is applied specifically to medical and biological QA, demonstrating its effectiveness across multiple specialized formats. Fine-tuned models achieve up to 10–30\% higher accuracy in closed-ended QA tasks and significantly enhance response quality in long-form and short-form QA.
\item A systematic evaluation of optimization strategies across closed-ended, long-form, and short-form QA provides a comprehensive analysis of biomedical AI applications. Fine-tuned models show notable improvements in ROUGE-1 scores for long-form QA, while RAG enhances short-form QA performance.
\item We leverage GPT-4o, the latest model, to assess its capabilities in fine-tuning, retrieval-augmented generation, and prompt engineering. Experimental results indicate that fine-tuning GPT-4o outperforms GPT-4 and GPT-3.5 across various biomedical QA benchmarks, underscoring the importance of domain adaptation for specialized applications.
\end{itemize}

By addressing challenges unique to the medical and biological fields—such as domain specificity, factual accuracy, and contextual depth—this research bridges the gap between general LLM optimization and the specialized needs of biomedical question answering. The results highlight the complementary benefits of fine-tuning and retrieval-based approaches, offering practical insights for developing AI-powered medical assistants and research tools.

\medskip

\section{Related Work}

\textbf{Optimizing LLMs for Medical and Biological Applications.} The application of LLMs in medical and biological domains has demonstrated significant improvements in reasoning-based question answering \cite{mcduff2023accuratedifferentialdiagnosislarge, Singhal2023}. One of the most notable advancements is Med-Gemini \citet{saab2024capabilitiesgeminimodelsmedicine}, a family of models fine-tuned specifically for medical reasoning. Med-Gemini has achieved state-of-the-art performance on the MedQA benchmark \citep{saab2024capabilitiesgeminimodelsmedicine}, surpassing previous models through an uncertainty-guided search strategy. This strategy enables the model to refine its responses based on external information retrieved through web search, ensuring greater factual accuracy and reliability.

Long-form question answering in medical and biological domains presents unique challenges, requiring models to generate coherent, factually accurate, and contextually rich responses. To address this, both Med-Gemini \citep{saab2024capabilitiesgeminimodelsmedicine} and OLAPH \citep{jeong2024olaphimprovingfactualitybiomedical} have adopted pairwise evaluation methodologies, allowing human experts to assess and compare generated answers against ground truth references. This approach ensures that models produce responses that align with expert consensus while minimizing factual errors and hallucinations.

OLAPH, on the other hand, employs a preference-based optimization framework to iteratively refine long-text generation. By constructing synthetic preference sets and training on preferred responses, OLAPH enhances factual consistency and linguistic fluency. In parallel, models such as BioGPT \cite{Zhang_2024} and Flan-PaLM \cite{Singhal2023} have explored domain-specific fine-tuning to improve the accuracy of medical text generation through task-specific retrieval mechanisms.

These methodologies underline the need for robust evaluation frameworks in medical LFQA. Given the complexity of clinical reasoning, future research should focus on refining automated evaluation metrics and incorporating multi-expert consensus validation to ensure reliability in real-world applications.

\textbf{Search-based Text Generation for Medical and Biological Tasks.}
RAG has emerged as a critical technique for enhancing the reliability and factual accuracy of LLM-generated responses in medicine. Med-Gemini integrates a search retrieval mechanism that dynamically incorporates external knowledge sources into its generation pipeline. This approach, known as uncertainty-guided search, enables the model to refine its answers based on retrieved web documents, reducing the risk of hallucinated content.

Beyond structured search, Med-Gemini’s multimodal capabilities allow cross-modal retrieval and reasoning, integrating text, images, and structured data from electronic health records. These capabilities make it particularly suitable for real-world applications such as clinical decision support, medical summarization, and literature synthesis.

Search-based text generation represents a promising direction for medical AI, enabling models to dynamically incorporate external knowledge while maintaining fluency and coherence.

\begin{figure}
\vskip 0.2in
\begin{center}
\centerline{\includegraphics[width=\columnwidth]{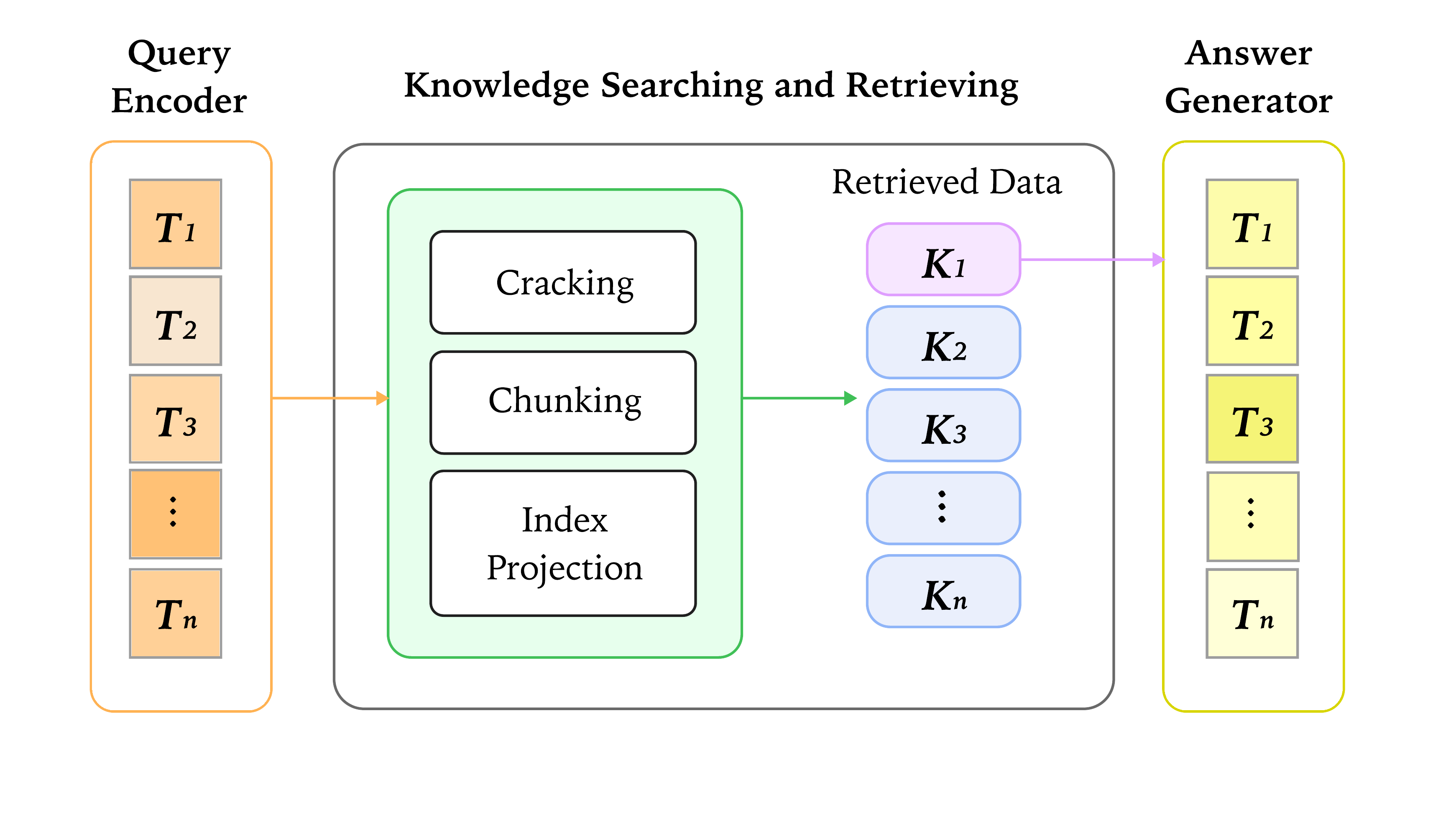}}
\caption{Illustration of the Retrieval-Augmented Generation (RAG) process. The system consists of three main components: (1) \textbf{Query Encoder}, which processes the input query into tokenized representations (\(T_1, T_2, \dots, T_n \)), (2) \textbf{Knowledge Searching and Retrieving}, where the system performs document cracking, chunking, and index projection to retrieve relevant knowledge (\(K_1, K_2, \dots, K_n\)), and (3) \textbf{Answer Generator}, which integrates retrieved data into the response generation process. Here, \(T_i\) represents tokenized query input, while \(K_i\) denotes retrieved knowledge chunks. This approach enhances factual accuracy by incorporating external knowledge into the model’s output.}
\label{RAG_mechanism}
\end{center}
\vskip -0.2in
\end{figure}

\section{Methodology}
\label{methodology}
To optimize LLMs for biomedical question-answering tasks, this study integrates fine-tuning, RAG, and prompt engineering (\cref{overview}). Each component plays a crucial role in enhancing different aspects of model performance: fine-tuning aligns the model with domain-specific knowledge, RAG dynamically retrieves relevant biomedical information to improve factual accuracy, and prompt engineering refines response generation based on task-specific requirements. Experiments were conducted in the Azure cloud environment, leveraging its scalable computing infrastructure to efficiently fine-tune models, and optimize inference performance \cite{MicrosoftAzureFineTuning2024, MicrosoftAzureAISearch2024}.

\subsection{Datasets for QA Tasks in Medicine and Biology}

To evaluate model performance in medical and biological question-answering tasks, multiple datasets covering closed-ended, long-form, and short-form QA formats were utilized. Detailed descriptions of the datasets utilized for fine-tuning can be found in \cref{exp-details}.

Close-ended question-answering datasets focus on multiple-choice or Boolean-style questions, where the model selects the most appropriate answer from a predefined set of options. MedQA \cite{jin2020diseasedoespatienthave} is derived from USMLE (United States Medical Licensing Examination), containing multiple-choice questions. PubMedQA (PQA-L) consists of biomedical research abstracts with questions that require binary or ternary reasoning-based answers (Yes, No, Maybe), testing a model’s ability to comprehend and infer conclusions from scientific literature \cite{Jin2019PubMedQA}. BioSQA is a structured biomedical QA dataset. It also contains 'Yes or No' type questions \cite{BioASQ}.

Long-form QA datasets are designed for generating comprehensive, well-structured answers instead of selecting from predefined choices. PubMedQA (PQA-A)'s reasoning labels were artificially generated. MedicationQA contains questions related to medications. LiveQA \cite{LiveMedQA2017} is a dynamic dataset based on real-world medical inquiries from users, evaluating a model’s ability to handle open-ended medical questions in diverse contexts. BioSQA \cite{BioASQ} focuses on biomedical and life sciences QA, requiring in-depth domain knowledge for accurate responses. We also created a custom dataset curated for training biomedical-specific language models, incorporating various QA formats from multiple sources to improve medical reasoning and factual accuracy.

Short-form QA datasets are designed to evaluate a model’s ability to generate concise yet accurate responses to medical questions. MedQA (short-form subset) is a refined version of the MedQA dataset, focusing on generating brief, direct answers to clinical case-based queries. Unlike long-form QA tasks, which require detailed explanations, short-form QA emphasizes precision, making it suitable for real-time applications such as clinical decision support and rapid medical reference systems.

\subsection{Fine-tuning LLMs}
To enhance LLMs for specialized medical question-answering, fine-tuning was performed using domain-specific datasets. The process involved supervised learning with labeled biomedical question-answer pairs, optimizing batch size, epochs, and learning rate for computational efficiency and model improvement. Adaptive optimization was applied, dynamically adjusting hyperparameters based on dataset complexity, while ensuring reproducibility through automatic seed assignment. Fine-tuned models demonstrated superior accuracy in structured reasoning tasks and improved response relevance in free-text generation, highlighting the effectiveness of task-specific adaptation for clinical and research applications. The details of the datasets and training configurations are summarized in \cref{exp-details}.

The default values were used for task parameters, allowing for adaptive optimization during training. The batch size was set to 0.2\% of the total training examples. The learning rate was determined based on the original pre-training rate, multiplied by a dynamic scaling factor, typically ranging between 0.5 and 2, ensuring an optimal balance between convergence speed and generalization. The number of training epochs was dynamically adjusted based on dataset size and complexity, allowing for effective learning without manual tuning. Additionally, the seed for randomization was automatically assigned to maintain reproducibility across training runs, ensuring consistency in fine-tuning results.

By leveraging fine-tuning, the model demonstrated improved performance in domain-specific evaluations, achieving higher accuracy in structured reasoning tasks and more relevant responses in free-text generation.

\subsection{Retrieval-Augmented Generation}
RAG was integrated to enhance medical question-answering by structuring and optimizing search efficiency. The overall RAG pipeline, including query encoding, document retrieval, and answer generation, is illustrated in \cref{RAG_mechanism}. This diagram highlights how queries (\(T_1, T_2, \dots, T_n\)) are processed, relevant knowledge chunks (\(K_1, K_2, \dots, K_n\)) are retrieved, and the final response is generated by integrating retrieved data.

To achieve this, a robust indexing framework was designed to systematically store and retrieve medical queries and their corresponding answers. The system was configured to support high-precision retrieval, ensuring that relevant information could be efficiently surfaced for downstream processing by language models. 

A keyword-based search strategy was adopted to enable deterministic and precise query matching. This method was chosen to facilitate structured data retrieval, ensuring that medical questions and answers were retrieved with minimal ambiguity. Unlike semantic or vector-based retrieval, the keyword approach provides consistent and interpretable results, making it particularly useful for retrieving well-structured medical data.

To further refine retrieval effectiveness, a structured field mapping strategy was implemented. The indexed fields included an ID field as a unique identifier, a question field for medical queries, and an answer field containing corresponding responses. The ID field was designated as the primary key to maintain data integrity, while both the question and answer fields were configured as searchable and retrievable. Additionally, text processing techniques were applied, including Microsoft’s English-language analyzer, to optimize tokenization and improve search relevance.

The system was designed to continuously update the indexed data through automated ingestion pipelines. These pipelines processed structured medical content, applied text analysis techniques, and periodically refreshed the search index.

\subsection{Prompting Strategies and Tuning Parameters}
Prompt engineering strategies have been expolored in medical and biomedical domains \cite{nori2023generalistfoundationmodelsoutcompete}. This study evaluates three QA setups—closed-ended, short-form, and long-form by modifying system prompts and decoding parameters. Closed-ended QA enforces strict output constraints with predefined options, using low temperature (0.1), top-p (0.7), and a frequency penalty (0.5) to ensure deterministic, high-confidence selections while preventing extraneous text generation. Long-form QA prioritizes detailed, structured responses for complex medical queries, increasing max tokens to 300 while maintaining temperature (0.2) and top-p (0.8) for scientifically accurate and coherent explanations. Unlike other setups, no penalties are applied, allowing for comprehensive literature-based reasoning and clinical insights. Short-form QA balances precision and informativeness, allowing up to 50 tokens with moderate temperature (0.2) and top-p (0.85) for concise yet accurate responses. No explicit stop sequence is enforced, and penalties are minimized to prevent unnecessary repetition or hallucination. These decoding strategies demonstrate that optimizing generation parameters based on answer format enhances medical QA effectiveness. Future work may explore dynamic prompt tuning techniques to further adapt the model’s responses based on real-world medical contexts.

\subsection{Evaluation Metrics}

For closed-domain question answering with reasoning, accuracy was the primary evaluation metric. When it comes to long-form and short-form question answering, where responses are more open-ended, a set of text generation evaluation metrics was employed. ROUGE (Recall-Oriented Understudy for Gisting Evaluation) was used to measure the lexical overlap between generated and reference responses. Specifically, ROUGE-1 captures unigram overlap, ROUGE-2 captures bigram overlap, and ROUGE-L measures the longest common subsequence (LCS), reflecting fluency and structural similarity. The ROUGE score is calculated as follows:
\begin{equation}
    \frac{\sum_{s \in \text{Reference}} \sum_{w \in s} \mathbb{I}(w \in \text{Generated})}
    {\sum_{s \in \text{Reference}} \sum_{w \in s} \mathbb{I}(w)}
\end{equation}
where \( w \) represents an n-gram in the reference or generated text, and \( \mathbb{I} \) is an indicator function.

In addition to ROUGE, BLEU (Bilingual Evaluation Understudy) was used to assess precision by measuring the overlap of n-grams between generated and reference responses. It is calculated as:
\begin{equation}
    BP \cdot \exp \left( \sum_{n=1}^{N} w_n \log p_n \right)
\end{equation}
where \( p_n \) represents the precision of n-grams, \( w_n \) are weighting factors, and \( BP \) is a brevity penalty to account for length mismatches.

Based on the evaluation methodology presented in OLAPH \cite{jeong2024olaphimprovingfactualitybiomedical}, we employed BERTScore and BLEURT (Bilingual Evaluation Understudy with Representations from Transformers) to evaluate semantic similarity beyond lexical overlap. BERTScore uses contextualized embeddings from a pre-trained transformer model to compute cosine similarity between reference and generated responses. BLEURT, a learned metric incorporating deep learning models, compares responses against human-written references to provide a more nuanced evaluation of response quality. These combined metrics offer a robust framework for assessing generated medical responses in terms of both lexical similarity and semantic relevance.

\section{Evaluation}

\subsection{Closed Question Answering with Reasoning }

Closed-domain question answering (QA) involves selecting an answer from a predefined set of options, such as multiple-choice formats (A, B, C, D) or Boolean-style questions (Yes, No, Maybe). We evaluate the performance of different models on three biomedical datasets—MedQA (multiple-choice medical board exam questions), PubMedQA (yes/no/maybe biomedical research questions), and BioASQ (yes/no biomedical factual questions)—to assess their reasoning ability in structured QA formats.

As shown in \cref{closed-qa-performance}, fine-tuned MedBioLM consistently achieves the highest accuracy across all datasets, demonstrating the benefits of domain-specific adaptation. In MedQA \cite{jin2020diseasedoespatienthave}, where models must select the most accurate answer from four options, MedBioLM achieves 88.0\% accuracy, outperforming all base models, including GPT-4o (87.0\%) and GPT-4 (81.71\%). Similarly, in PubMedQA, which requires interpreting research abstracts to determine the correctness of a biomedical hypothesis, MedBioLM (78.9\%) surpasses all other models, significantly outperforming the GPT-4o base model.

For BioASQ, a factual biomedical QA dataset with Yes/No answers, MedBioLM achieves near-perfect accuracy (96\%), comparable to GPT-4 (96\%) and outperforming GPT-4o (92\%). This suggests that while general-purpose large language models (LLMs) like GPT-4o provide strong baseline accuracy, fine-tuning significantly improves performance on more complex reasoning tasks like PubMedQA, where domain knowledge and inference are essential.

The results highlight the advantages of fine-tuning in closed-domain biomedical QA, particularly for datasets that require research-based reasoning (e.g., PubMedQA). While general-purpose LLMs perform well in simple factual QA (e.g., BioASQ), domain-adapted models like MedBioLM provide a clear advantage in specialized, research-intensive medical reasoning tasks.

The results in \cref{response-distribution} highlight a key challenge in reasoning tasks within PubMedQA and BioASQ: models with lower accuracy tend to produce a significantly higher proportion of “Maybe” responses, indicating uncertainty or lack of confidence in their answers. For example, GPT-3.5, which achieved the lowest accuracy in PubMedQA (19\%), selected “Maybe” for 82\% of the questions, suggesting a tendency to avoid committing to definitive answers. These findings suggest that fine-tuning on domain-specific datasets helps improve model confidence and correctness in biomedical reasoning tasks, reducing indecisive outputs and improving decision-making reliability.

\begin{table*}[t]
\centering
\caption{Performance comparison of models on Closed-Domain Question Answering (Multiple-Choice and Boolean QA). The table presents accuracy scores across different datasets for MedBioLM, GPT-4o, GPT-4o-mini, GPT-4, and GPT-3.5.MedBioLM consistently outperforms general-purpose models, achieving the highest accuracy in MedQA, PubMedQA, and BioASQ}
\label{closed-qa-performance}
\vspace{0.4cm} 
\begin{small}
\begin{tabular}{l r r r r r}
\toprule
\textbf{Dataset} & \textbf{MedBioLM} & \textbf{GPT-4o} & \textbf{GPT-4o-mini} & \textbf{GPT-4} & \textbf{GPT-3.5} \\
\midrule
\textbf{MedQA} & \textbf{88.0} & 87.0 & 70.4 & 81.71 & 50.51 \\
\textbf{PubMedQA} & \textbf{78.9} & 44.74 & 77.55 & 70.0 & 19.30 \\
\textbf{BioASQ} & \textbf{96.0} & 92.0 & 92.0 & 96.0 & 88.0 \\
\bottomrule
\end{tabular}
\end{small}
\end{table*}

\begin{table*}[t]
\centering
\caption{Response distribution and accuracy of different models on PubMedQA and BioASQ reasoning tasks. The table presents the percentage of responses classified as "Yes," "No," and "Maybe" for PubMedQA and "Yes" and "No" for BioASQ, along with overall accuracy scores for each dataset.}
\label{response-distribution}
\vspace{0.3cm} 
\begin{small}
\begin{tabular}{l r r r r r r r}
\toprule
 & \multicolumn{3}{c}{\textbf{PubMedQA}} & \textbf{Accuracy} & \multicolumn{2}{c}{\textbf{BioASQ}} & \textbf{Accuracy} \\
\cmidrule(lr){2-4} \cmidrule(lr){6-7}
\textbf{Model} & \textbf{Yes (\%)} & \textbf{No (\%)} & \textbf{Maybe (\%)} &  & \textbf{Yes (\%)} & \textbf{No (\%)} &  \\
\midrule
\textbf{MedBioLM} & 68.42 & 12.28 & 19.30 & \textbf{78.9} & 64 & 36 & \textbf{96} \\
\textbf{GPT-4o} & 39.47 & 1.75 & 58.77 & 44.74 & 68 & 32 & 92 \\
\textbf{GPT-4o-mini} & 75.51 & 6.12 & 18.37 & 77.55 & 60 & 40 & 92 \\
\textbf{GPT-4} & 68.00 & 2.00 & 30.00 & 70 & 64 & 36 & 96 \\
\textbf{GPT-3.5} & 14.04 & 4.39 & 81.58 & 19.30 & 72 & 28 & 88 \\
\bottomrule
\end{tabular}
\end{small}
\end{table*}

\subsection{Long-form Question Answering }

Long-form question evaluation results (\cref{lf-perfromance}) demonstrate that fine-tuning generally enhances model performance, though the degree of improvement varies across datasets. In the MedicationQA dataset, the fine-tuned GPT-4o model significantly outperforms the base model across all metrics. ROUGE-1 increases from 19.85 to 24.69, ROUGE-2 from 4.20 to 8.80, and BLEU more than doubles from 0.98 to 2.49. Additionally, BERTScore improves from -7.63 to 8.98, indicating a substantial enhancement in semantic alignment with human-written responses. These findings suggest that fine-tuning is particularly effective for medication-related long-form question-answering tasks.

In contrast, the results for the LiveQA dataset present a more complex trend. The fine-tuned GPT-4o model performs slightly worse than the base model in ROUGE-1 (24.12 vs. 26.96) and ROUGE-L (13.31 vs. 13.42), suggesting that fine-tuning may have introduced some degree of overfitting, reducing the model’s ability to generalize to unseen test data. However, improvements in ROUGE-2 (6.18 vs. 5.80) and BLEU (1.63 vs. 1.41) indicate enhanced phrase-level fluency. Further analysis is necessary to determine whether this discrepancy is due to the characteristics of the training data or specific nuances in the LiveQA dataset.

\begin{figure}
\vskip 0.2in
\begin{center}
\centerline{\includegraphics[width=\columnwidth]{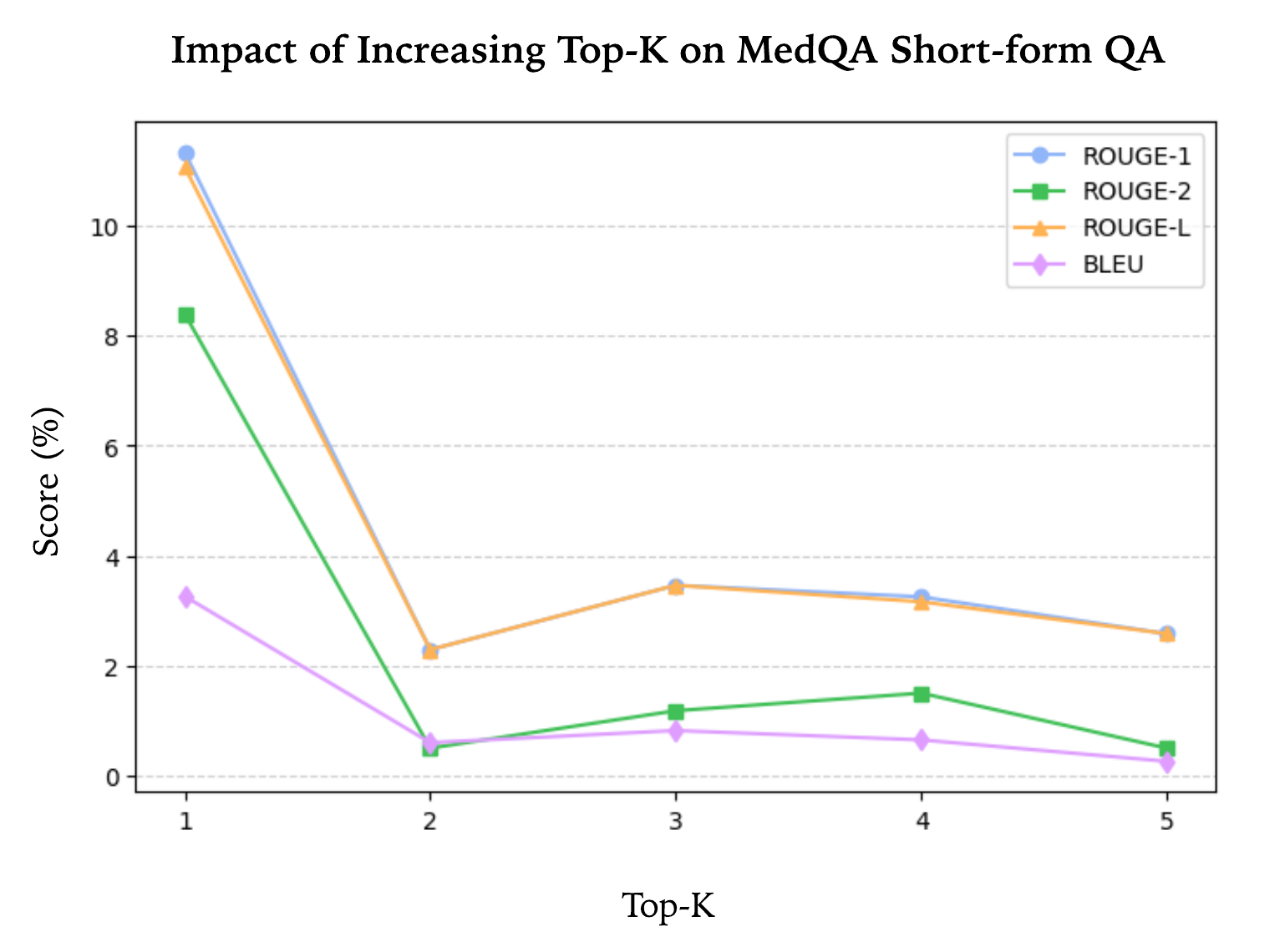}}
\caption{Impact of increasing Top-K on MedQA short-form QA. As the number of retrieved documents increases, the performance of all evaluation metrics decreases. Given the nature of the task, which expects concise short-form answers, retrieving more documents introduces noise and conflicting information, negatively affecting answer quality.}
\label{top_k}
\end{center}
\vskip -0.2in
\end{figure}
The fine-tuned MedBioLM model, trained exclusively for long-form question answering across multiple medical datasets (LiveQA, MedicationQA, PubMedQA, and BioASQ), achieves the highest performance across most metrics. It attains the best ROUGE-1 (26.67), BLEU (3.12), and ROUGE-L (18.71) scores, suggesting that training on a diverse range of datasets enhances the model’s ability to generalize across different biomedical question-answering tasks. Additionally, BERTScore (12.08) is the highest among all models, indicating stronger alignment with reference answers. BLEURT, while still negative (-30.26), is relatively better than in single-dataset fine-tuned models.

A major challenge across all models is the consistently negative BLEURT scores, suggesting that none of the models produce outputs that closely resemble human-written responses. While fine-tuning does improve BLEURT in certain cases, particularly for MedicationQA (-33.82) and MedBioLM (-30.26), these results highlight the need for further advancements in generating more natural and human-like long-form answers. Future research should explore additional methods, such as retrieval-augmented generation and reinforcement learning from human feedback, to further refine the performance of long-form medical QA models.

\begin{table*}[t]
\centering
\caption{Long-form Question Answering: Comparative performance of fine-tuned and base models on different datasets. The table presents ROUGE, BLEU, BERTScore, and BLEURT scores for LiveQA, MedicationQA, and a combined dataset. The Custom Dataset consists of a combination of long-form QA data from LiveQA, MedQA, PubMedQA, and BioASQ.}
\label{lf-perfromance}
\vspace{0.3cm} 

\centerline{\makebox[0.9\textwidth]{
\begin{tabular}{l l r r r r r r}
\toprule
\textbf{Model} & \textbf{Dataset} & \textbf{ROUGE-1} & \textbf{ROUGE-2} & \textbf{ROUGE-L} & \textbf{BLEU} & \textbf{BERTScore} & \textbf{BLEURT} \\
\midrule
Fine-Tuned GPT-4o & LiveQA & 24.12 & 6.18 & 13.31 & 1.63 & 1.10 & -46.48 \\
GPT-4o Base Model & LiveQA & \textbf{26.96} & 5.80 & 13.42 & 1.41 & -2.93 & -34.79 \\
Fine-Tuned GPT-4o & MedicationQA & 24.69 & \textbf{8.80} & 17.61 & 2.49 & 8.98 & -33.82 \\
GPT-4o Base Model & MedicationQA & 19.85 & 4.20 & 10.97 & 0.98 & -7.63 & -33.21 \\
 MedBioLM  & Custom Dataset & 26.67 & 8.72 & \textbf{18.71} & \textbf{3.12} & \textbf{12.08} & \textbf{-30.26} \\
\bottomrule
\end{tabular}
}}
\end{table*}

\begin{figure}
\vskip 0.2in
\begin{center}
\centerline{\includegraphics[width=\columnwidth]{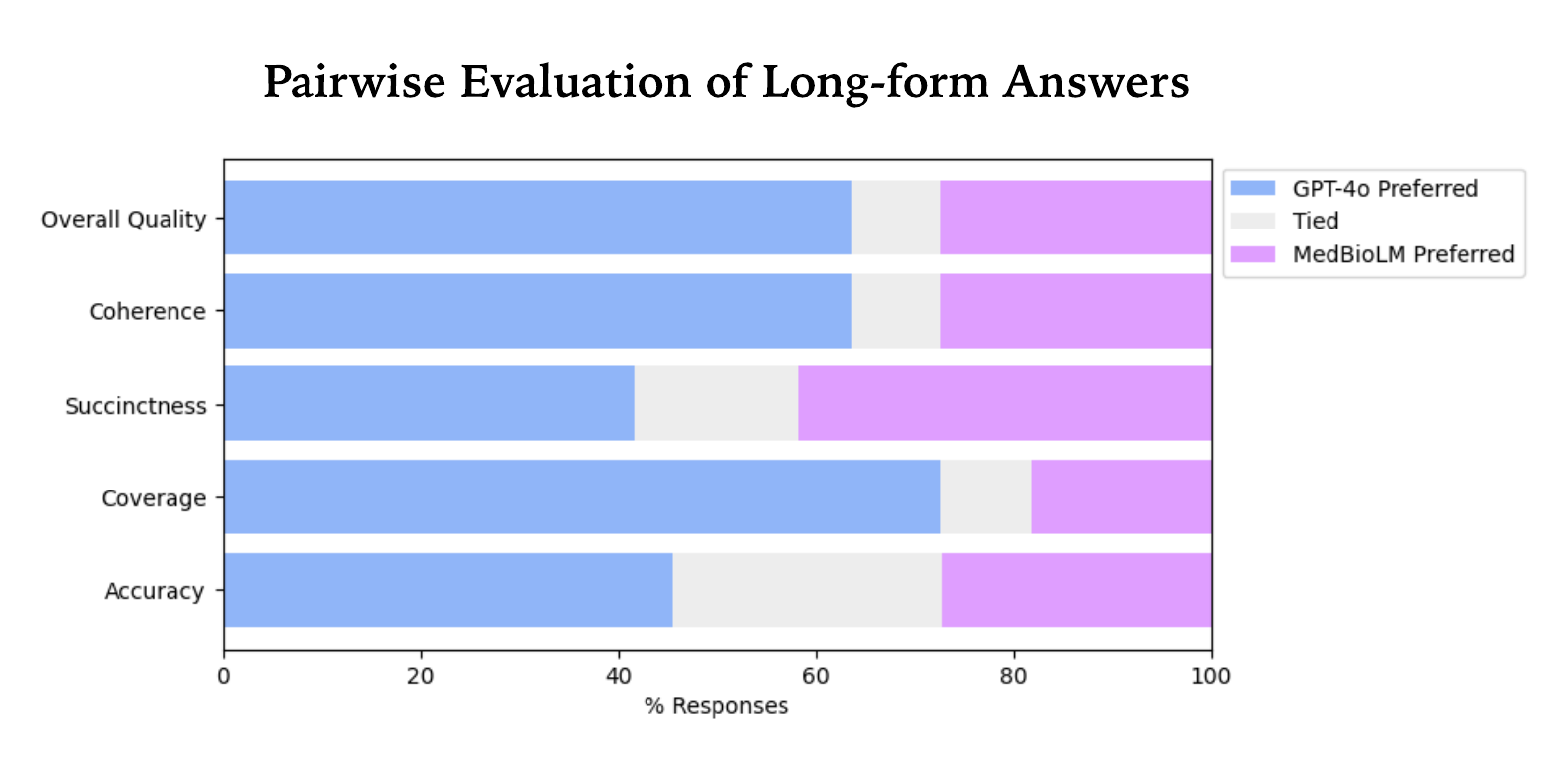}}
\caption{Pairwise evaluation of long-form answers comparing GPT-4o and MedBioLM across five key criteria: overall quality, coherence, succinctness, coverage, and accuracy. Bars represent the percentage of responses where GPT-4o was preferred (blue), MedBioLM was preferred (purple), or the responses were rated as tied (gray).}
\label{pairwise_evaluation}
\end{center}
\vskip -0.2in
\end{figure}

\begin{table*}
\caption{Comparative Performance of Different Models on MedQA Short-form Dataset. The table presents ROUGE, BLEU, BERTScore, and BLEURT scores for different configurations of GPT-4o and MedBioLM models, with and without RAG.}
\label{shortform-eval}
\vspace{0.3cm} 

\centering
\begin{tabular}{l r r r r r r}
\toprule
\textbf{Model} & \textbf{ROUGE-1} & \textbf{ROUGE-2} & \textbf{ROUGE-L} & \textbf{BLEU} & \textbf{BERTScore} & \textbf{BLEURT} \\
\midrule
GPT-4o           & 4.35  & 1.14  & 4.22  & 0.28  & -20.45  & -141.61  \\
GPT-4o + RAG     & 6.86  & 2.42  & 6.33  & 0.34  & -20.71  & -70.49  \\
MedBioLM         & \textbf{43.17}  & 25.65  & \textbf{42.73}  & \textbf{11.99}  & \textbf{39.16}  & \textbf{33.43}  \\
MedBioLM + RAG   & 40.51  & \textbf{25.79}  & 40.19  & 11.55  & 33.43  & 8.63  \\
\bottomrule
\end{tabular}
\end{table*}

\textbf{Qualitative Observation.} Pair-wise evaluation \cite{saab2024capabilitiesgeminimodelsmedicine, jeong2024olaphimprovingfactualitybiomedical} was used for qualitative observation of long-form QA. Pair-wise evaluation is a method used to compare two models by directly assessing their outputs against each other for the same set of questions. In this analysis, a single general physician conducted a pair-wise evaluation of GPT-4o and MedBioLM across five key metrics: accuracy, coverage, succinctness, coherence, and overall quality. The evaluation criteria were defined as follows: accuracy measured whether the response provided correct and relevant information, coverage assessed whether the response sufficiently addressed all relevant aspects of the question, succinctness evaluated whether the response was concise without unnecessary length, coherence examined the logical flow and readability of the response, and overall quality reflected the evaluator’s holistic judgment, incorporating both objective and subjective aspects of the response.

In \cref{pairwise_evaluation}, we show findings suggesting GPT-4o generally provides more accurate and comprehensive responses, while MedBioLM demonstrates comparable performance in certain aspects, such as succinctness and coherence. However, in some metrics, the performance gap between the two models was relatively small, and a notable proportion of responses were tied, particularly in succinctness.

It is important to note that this evaluation was conducted by a single physician, which may introduce subjectivity in the assessment. Since this analysis is part of a broader study, future research will involve collaboration with medical professionals from various specialties to ensure a more objective and comprehensive evaluation. Through this approach, the goal is to further enhance the performance of the MedBioLM model and refine it into a more robust medical QA system.

\subsection{Short-form Question Answering}

The short-form question evaluation results (\cref{shortform-eval}) indicate that the fine-tuned GPT-4o model, both with and without RAG, substantially outperforms the base model across all key evaluation metrics.	The short-form question evaluation results indicate that the fine-tuned GPT-4o model, with or without RAG, significantly outperforms the base model across all key evaluation metrics. The ROUGE-1 score improves substantially from 4.35 in the base model to 43.17 in the fine-tuned model, demonstrating a notable enhancement in answer relevance and quality. Similarly, BLEU scores rise from 0.28 in the base model to 11.55 in the fine-tuned model, indicating substantial improvement in fluency and syntactic accuracy. Furthermore, BERTScore, which measures semantic alignment with reference answers, shows a dramatic improvement from -20.45 in the base model to 33.43 in the fine-tuned model, highlighting the effectiveness of fine-tuning in enhancing content coherence.

While RAG provides marginal improvements for the base model, it does not bridge the substantial performance gap with the fine-tuned model. For example, the ROUGE-1 score for the Base Model + RAG rises to 6.86, a slight increase from 4.35 in the base model alone, but still far below the 43.17 achieved by the fine-tuned model. These findings suggest that retrieval augmentation alone is insufficient to significantly enhance performance unless combined with fine-tuning.

BLEURT scores exhibit the most significant performance gap. The base model records a BLEURT score of -141.61, indicating poor semantic similarity with reference answers. In contrast, the fine-tuned GPT-4o model achieves a BLEURT score of 8.63, demonstrating that fine-tuning significantly improves the model’s ability to generate responses that closely align with human-written answers.

Notably, there is no substantial difference between the fine-tuned GPT-4o model and its RAG-augmented counterpart. ROUGE and BLEU scores remain nearly identical across these configurations, suggesting that RAG does not contribute substantial performance gains once fine-tuning has been applied. This underscores the dominant role of fine-tuning in improving model performance, while RAG provides only minor effects to an already well-trained model.

MedQA short-form QA results show that increasing top-k (the number of retrieved documents) does not necessarily improve performance (\cref{top_k}). The best results were observed at k=1, where ROUGE-1 reached 11.33 and BLEU was 3.26. As top-k increased, all evaluation metrics declined, with ROUGE-1 dropping to 2.58 at k=5. These findings indicate that retrieving more documents introduces noise and conflicting information, reducing overall answer quality. The model struggles to synthesize relevant content effectively, leading to lower lexical similarity with reference answers. This suggests that for closed-domain QA tasks, retrieval strategies should prioritize precision over recall, and an optimal top-k value should be carefully selected to balance informativeness and accuracy.

\section{Conclusion}
In this work, we explore the optimization of LLMs for biomedical QA by integrating fine-tuning, RAG, and prompt engineering. Through experiments using a state-of-the-art LLM as the base model, we developed a domain-specific approach tailored to biomedical QA. Our findings demonstrate that fine-tuning significantly enhances structured reasoning in closed-ended QA tasks, while prompt engineering plays a crucial role in optimizing response clarity and coherence for both short-form and long-form biomedical answers. These results highlight the importance of domain adaptation in improving LLM performance for specialized fields. 

However, challenges remain. RAG’s impact on factual accuracy was inconsistent, and fine-tuned models risk overfitting to specific datasets. Additionally, evaluation by a single medical professional may introduce bias. Future research should involve multiple domain experts, explore hybrid retrieval techniques, and leverage human-in-the-loop evaluation to enhance model reliability. This study advances biomedical AI, bridging the gap between general-purpose LLMs and real-world medical applications.

\nocite{langley00}

\bibliography{example_paper}
\bibliographystyle{icml2025}


\newpage
\appendix
\onecolumn
\section{Experimental Details}

\medskip
\medskip

\begin{table}[H]  
\caption{Experimental details of the fine-tuning process for various biomedical question-answering tasks. The table outlines the dataset used, training duration, base model, number of epochs, and batch size for each task. Fine-tuning was performed across closed-ended, long-form, and short-form QA datasets to optimize model performance. All base models used for fine-tuning were GPT-4o.}
\label{exp-details}
\vspace{0.3cm} 
\centering
\begin{small}
\begin{tabular}{l l r r r r}
\toprule
\textbf{Task} & \textbf{Train Dataset} & \textbf{Train Samples} & \textbf{Training Duration} & \textbf{Epochs} & \textbf{Batch Size} \\
\midrule
\multirow{3}{*}{Closed-ended QA} 
 & MedQA & 10,178 & 3h 25m 33s & 2 & 13 \\
 & PubMedQA (PQA-L) & 552 & 7h 46m 44s & 3 & 1 \\
 & BioSQA & 5,049 & 3h 10m 7s & 3 & 2 \\
\midrule
\multirow{5}{*}{Long-form QA} 
 & PubMedQA (PQA-A) & 196,144 & 1d 6h 29m 20s & 1 & 64 \\
 & MedicationQA & 551 & 1h 44m 18s & 3 & 1 \\
 & LiveQA & 500 & 1h 46m 17s & 3 & 1 \\
 & BioSQA & 5,049 & 2h 36m 49s & 3 & 10 \\
 & Combined Custom Dataset & 6,652 & 2h 6m 1s & 3 & 13 \\
\midrule
Short-form QA & MedQA & 10,178 & 1h 49m 44s & 2 & 13 \\
\bottomrule
\end{tabular}
\end{small}
\end{table}

\medskip

\medskip

\begin{table}[H]  
\caption{Comparison of prompting strategies and decoding parameters for Closed-Ended, Long-Form, and Short-Form Question Answering. The table outlines system messages, token limits, and decoding parameters optimized for different QA formats.}
\label{decoding-params}
\vspace{0.3cm} 
\centering
\begin{small}
\begin{tabularx}{\textwidth}{l X X X}
\toprule
\textbf{Parameter} & \textbf{Closed-Ended QA} & \textbf{Long-Form QA} & \textbf{Short-Form QA} \\
\midrule
\textbf{System Message} & 
"You are an expert medical AI assistant. Answer the following question using only one letter: A, B, C, or D." & 
"You are a biomedical research expert. Generate precise and well-structured answers." & 
"You are an expert medical AI assistant. Provide concise and accurate answers." \\
\midrule
\textbf{Max Tokens} & 2 & 300 & 50 \\
\textbf{Temperature} & 0.1 & 0.2 & 0.2 \\
\textbf{Top P} & 0.7 & 0.8 & 0.85 \\
\textbf{Frequency Penalty} & 0.5 & 0.0 & 0.2 \\
\textbf{Presence Penalty} & 0.1 & 0.0 & 0.0 \\
\textbf{Stop Sequence} & ["\textbackslash n"] & - & - \\
\bottomrule
\end{tabularx}
\end{small}
\end{table}

\medskip

\cref{decoding-params} summarizes the prompting strategies and decoding parameters optimized for different types of biomedical QA tasks. It highlights variations in system messages, token limits, and decoding parameters tailored for closed-ended, long-form, and short-form question answering to optimize accuracy and response quality.

\begin{figure}[ht]
\vskip 0.2in
\begin{center}
\centerline{\includegraphics[width=0.9\columnwidth]{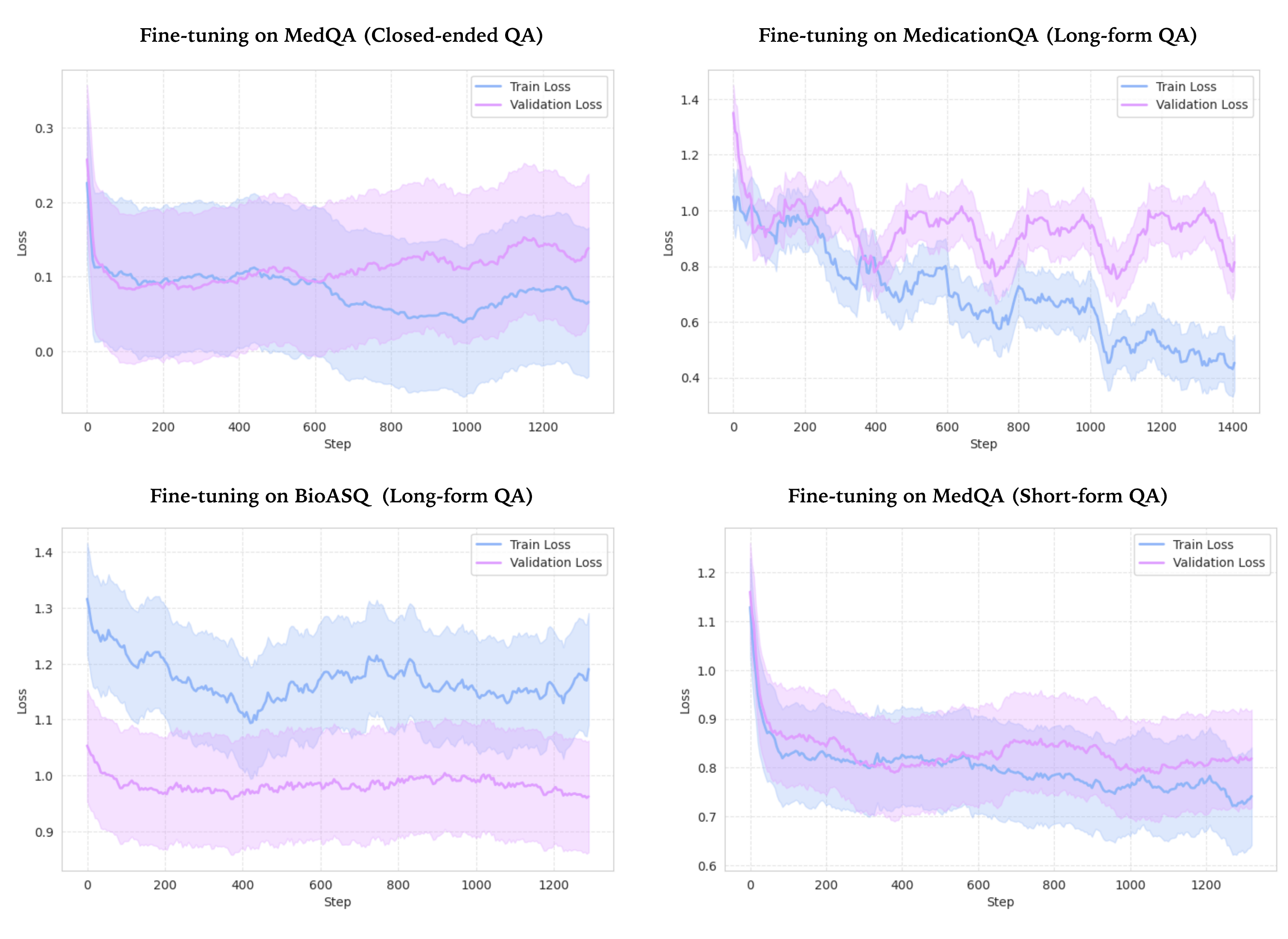}}
\caption{Fine-tuning loss curves for different biomedical QA tasks. Each plot illustrates the training loss (blue) and validation loss (purple) over training steps. The shaded regions represent the standard deviation of loss values. The four subplots correspond to different datasets and QA formats: (Top-left) MedQA (Closed-ended QA), (Top-right) MedicationQA (Long-form QA), (Bottom-left) BioASQ (Long-form QA), and (Bottom-right) MedQA (Short-form QA). The decreasing trend in loss indicates effective model adaptation, though variance in validation loss highlights differences in dataset complexity and generalization performance.}
\label{training_logs}
\end{center}
\vskip -0.2in
\end{figure}

\end{document}